\documentclass[conference]{IEEEtran}
\usepackage{latexsym}
\usepackage{array}
\usepackage{amsthm}
\usepackage{float}
\usepackage{mathtools}
\usepackage{amssymb}
\usepackage{wrapfig}
\usepackage[dvipsnames]{xcolor}

\usepackage{cite}
\usepackage{amsmath,amssymb,amsfonts}
\usepackage{multirow, threeparttable}
\usepackage{algorithmic}
\usepackage{graphicx}
\usepackage{textcomp}
\usepackage{xcolor}\usepackage{tablefootnote}
\usepackage[a4paper, total={184mm,244mm}]{geometry}
\def\BibTeX{{\rm B\kern-.05em{\sc i\kern-.025em b}\kern-.08em
    T\kern-.1667em\lower.7ex\hbox{E}\kern-.125emX}}

\usepackage{tikz}
\usetikzlibrary{backgrounds}
\usetikzlibrary{arrows,shapes}
\usetikzlibrary{tikzmark} 
\usetikzlibrary{calc} 

\usepackage[small,compact]{titlesec} 
\usepackage[compact]{titlesec}
\titlespacing{\section}{2pt}{2pt}{2pt}
\titlespacing{\subsection}{2pt}{2pt}{2pt}
\titlespacing{\subsubsection}{2pt}{2pt}{2pt}

\begin{document}
\newcommand{\highlight}[2]{\colorbox{#1!17}{$#2$}}
\newcommand{\highlightdark}[2]{\colorbox{#1!47}{$#2$}}

\title{Computational and Storage Efficient Quadratic Neurons for Deep Neural Networks}


\author{\IEEEauthorblockN{Chuangtao Chen\textsuperscript{1}, Grace Li Zhang\textsuperscript{2}, Xunzhao Yin\textsuperscript{3}, Cheng Zhuo\textsuperscript{3}, Ulf Schlichtmann\textsuperscript{1}, Bing Li\textsuperscript{1}}
\IEEEauthorblockA{\textsuperscript{1}Technical University of Munich, \textsuperscript{2}Technical University of Darmstadt, \textsuperscript{3}Zhejiang University
\\
\{chuangtao.chen, ulf.schlichtmann, b.li\}@tum.de, grace.zhang@tu-darmstadt.de, \{xzyin1, czhuo\}@zju.edu.cn}
}

\maketitle


\begin{abstract}
Deep neural networks (DNNs) have been widely deployed across diverse domains such as computer vision and natural language processing. However, the impressive accomplishments of DNNs have been realized alongside extensive computational demands, thereby impeding their applicability on resource-constrained devices. To address this challenge, many researchers have been focusing on basic neuron structures, the fundamental building blocks of neural networks, to alleviate the computational and storage cost. In this work, an efficient quadratic neuron architecture distinguished by its enhanced utilization of second-order computational information is introduced. By virtue of their better expressivity, DNNs employing the proposed quadratic neurons can attain similar accuracy with fewer neurons and computational cost. Experimental results have demonstrated that the proposed quadratic neuron structure exhibits superior computational and storage efficiency across various tasks when compared with both linear and non-linear neurons in prior work.
\end{abstract}



\section{Introduction} \label{sec:introduction}

Over the past decade, there has been an unprecedented success in the field of artificial neural networks. Neural networks, particularly those with deep layers, have been utilized in various tasks, including but not limited to, image classification \cite{alexnet,dosovitskiy2020image}, natural language processing \cite{transformer,gpt3} and generative tasks \cite{dalle2}. The excellent performance of deep neural networks (DNNs) are achieved at the cost of heavy computational and storage cost. These challenges have become a critical concern for devices with constrained computational and storage resources.

To mitigate the computational and storage burden of DNNs, there have been various research efforts including novel efficient network architectures \cite{tan2019efficientnet}, network pruning \cite{9116367,9474235,Petri2023,mengnan2023} and quantization \cite{9474037, 8806822,wenhao2023}. Recently, some researchers have been focusing on structures of individual neurons, which map multiple inputs to a single output followed by an activation function. To increase the efficiency of neurons, several studies have explored the use of more sophisticated non-linear neurons \cite{kervolutional_nn,9506221}, particularly second-order neurons, which are also referred to as quadratic neurons \cite{mantini2021cqnn,zoumpourlis2017non}. However, the parameter and computation costs associated with such non-linear neurons often make them unsuitable for deployment in practice. To reduce the parameter and computation overhead, some researchers have attempted to use quadratic neurons only in the initial or certain layers \cite{kervolutional_nn,zoumpourlis2017non,jiang2020nonlinear}, or to optimize complex neurons into more manageable forms \cite{fan2021expressivity,xu2020efficient,jiang2020nonlinear}. Nevertheless, there are still a large number of parameters and thus a high computation cost incurred by such quadratic neurons. In addition, the second-order information of neurons in these methods has not been exploited sufficiently, resulting in inefficient neuron implementations.

In this paper, we propose a novel efficient quadratic neuron for DNNs to address the challenges mentioned above. The proposed quadratic neuron possesses the following features:

\begin{itemize}
    \item By following a rigorous mathematical procedure rather than using a heuristic approach, we simplify quadratic neurons to reduce the incurred parameter and computation cost while preserving the expressivity. 
    \item To maximally utilize the second-order information of quadratic neurons to enhance the expressivity of DNNs, the intermediate results in the second-order computation are used as the partial output of the quadratic neuron. 
    \item The efficacy of the proposed neuron structure is evaluated in various tasks, such as convolutional neural networks (CNNs) and Transformers. Compared with state-of-the-art quadratic neurons \cite{fan2021expressivity,QuadraLib_xu}, neural networks with our neurons achieve a better or similar accuracy on CIFAR-10 dataset \cite{CIFAR_dataset} but with at least $24\%$ fewer parameters and $24\%$ less computational cost. In Transformer architecture, networks equipped with our proposed neuron structure can reduce both the computational cost and storage cost by more than $20\%$ with a better accuracy in the WMT14 German-English translation task.
\end{itemize}

This paper is structured as follows: Sec.~\ref{sec:motivation} provides a more in-depth overview of non-linear neurons, encompassing their benefits as well as their limitations. In Sec.~\ref{sec:method}, we present the construction of our proposed neuron, including a systematic simplifying procedure from a complex form. The effectiveness of our approach is evaluated by comparing the proposed neurons with the linear and other non-linear designs, and the experimental results as well as an in-depth analysis of the proposed quadratic neuron are reported in Sec.~\ref{sec:experiment}. Finally, Sec.~\ref{sec:conclusion} summarizes and concludes this paper.

\section{Background and Motivation} \label{sec:motivation}

This section presents the background and motivation of our work, highlighting various types of neurons explored by previous research, along with their respective advantages and drawbacks. We denote vectors with boldface lowercase letters and matrices with boldface uppercase letters. Scalars are represented by non-boldface lowercase letters. Unless otherwise stated, all vectors are assumed to be column vectors of length $n$ and matrices have dimensions of $n\times n$. Subscripts are used for indexing elements in a vector or a matrix. For example, $\mathbf{x}_3$ represents the third element (scalar) in vector $\mathbf{x}$.

\begin{figure}[tbp]
\centerline{\includegraphics[width=3.0in]{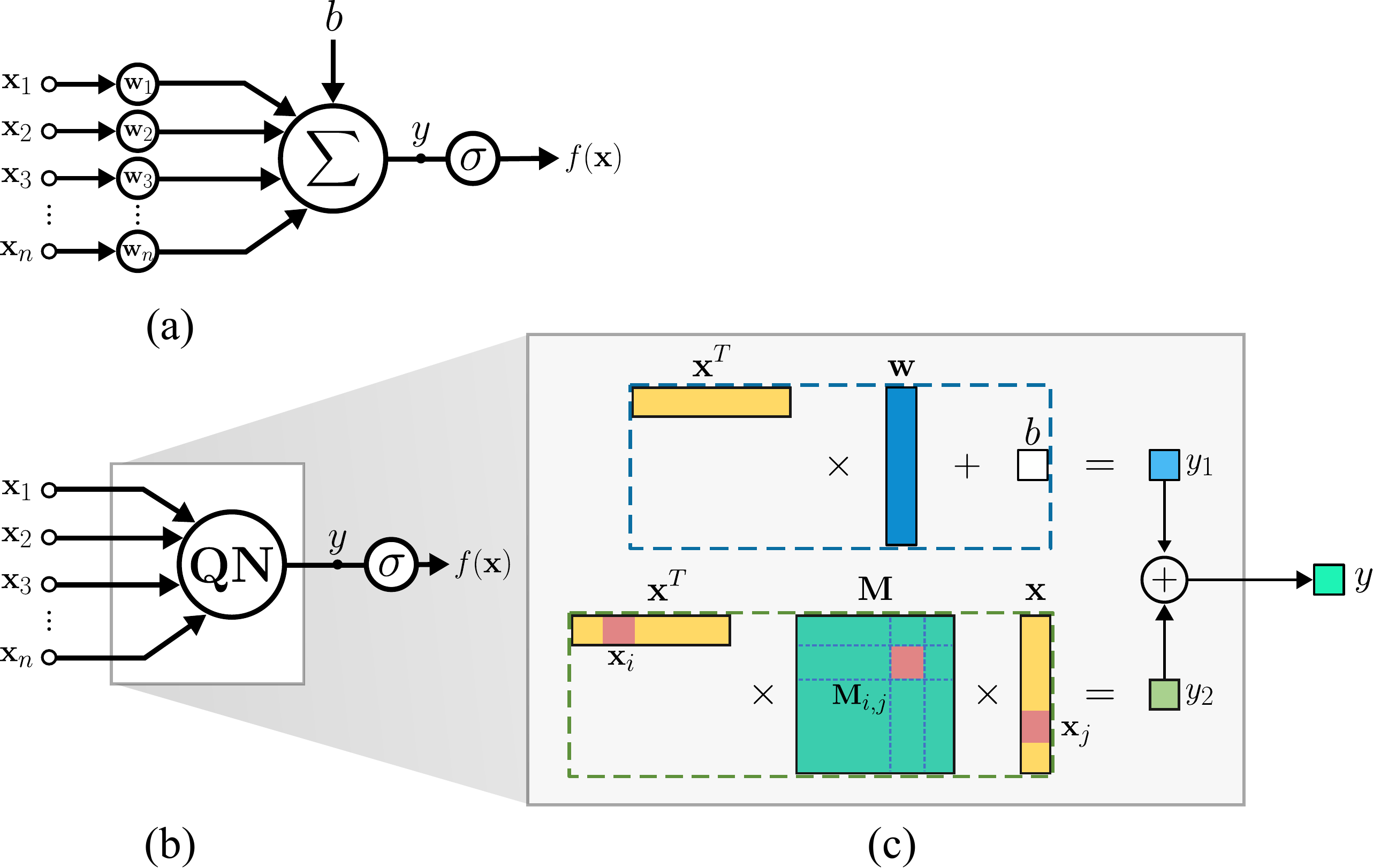}}
\caption{Structure of (a) the linear neuron and (b) the general quadratic neuron; (c) The computation process of the general quadratic neuron.} \label{fig:general_quad_neuron}
\end{figure}

\subsection{General Quadratic Neuron}
A neuron can be perceived as a multivariate function of its inputs. We denote the conventional linear neuron, which is shown in Fig.~\ref{fig:general_quad_neuron}a, as $f(\mathbf{x})= \sigma\left(\mathbf{w}^T\mathbf{x} + b\right)$, where $\mathbf{w}$ is the weight vector, $\mathbf{x}$ is the input vector, $b$ is the bias and $\sigma(\cdot)$ is the activation function. A linear neuron without the activation function can be viewed as a first-order polynomial function with respect to all the inputs with the bias serving as the zeroth-order term. In this paper, we only focus on the computation before the activation function in a neuron and the computational result is denoted as $y$ for simplicity.

To enhance the expressivity of neural networks, a typical approach is to replace linear neurons with non-linear neurons, such as the quadratic neuron \cite{mantini2021cqnn,zoumpourlis2017non,fan2021expressivity,QuadraLib_xu}. The general quadratic neuron can be represented mathematically as:
\begin{equation}\label{eq:gen_quad_neuron}
    y = 
        \tikzmarknode{p1}{\highlight{NavyBlue}{\mathbf{x}^T\mathbf{Mx}}}
        +
        \tikzmarknode{p2}{\highlight{WildStrawberry}{\mathbf{w}^T\mathbf{x} + b}}. 
\end{equation}
\begin{tikzpicture}[overlay,remember picture,>=stealth,nodes={align=left,inner ysep=1pt},<-]
        \node[anchor=east,color=NavyBlue!85,yshift=-0.8em,xshift=0em] (Istext) at (p1.south)
            {\textsf{\footnotesize quadratic part: $y_2$}};
        \draw [color=NavyBlue](p1.south) |- ([xshift=-0.3ex,yshift=-0.2ex]Istext.south west);
        
        \node[anchor=west,color=WildStrawberry!85,yshift=-0.8em,xshift=0em] (Istext) at (p2.south)
            {\textsf{\footnotesize linear part: $y_1$}};
        \draw [color=WildStrawberry](p2.south) |- ([xshift=-0.3ex,yshift=-0.2ex]Istext.south east);
\end{tikzpicture}

The general quadratic neuron in Eq.~\eqref{eq:gen_quad_neuron} is the most complex one and possesses the largest function space among all quadratic neurons. Fig.~\ref{fig:general_quad_neuron}b depicts the structure of a general quadratic neuron, while its computation process is shown in Fig.~\ref{fig:general_quad_neuron}c. As previously mentioned, the linear neuron on the left performs a summation of all weighted inputs and a bias term. In contrast, the computation of the general quadratic neuron is considerably more complex. Besides the parameters that also exist in a linear neuron, there is an additional parameter matrix $\mathbf{M}$ with dimensions of $n\times n$ in the general quadratic neuron. The final output of the general quadratic neuron $y$ is the sum of the linear term and the quadratic term. In Fig.~\ref{fig:general_quad_neuron}c we divide the computation in the quadratic neuron into two sections enclosed by dotted-line rectangles: the upper section represents the linear part comprising of the same computation as in a linear neuron, where the result is denoted as $y_1 = \mathbf{w}^T\mathbf{x} + b$. The lower section represents the quadratic part, where the element at the $i^{th}$ row and $j^{th}$ column of the matrix corresponds to the coefficient of the cross term $\mathbf{x}_i\mathbf{x}_j$ (depicted in red in Fig.~\ref{fig:general_quad_neuron}c). The quadratic term, represented by the equation $y_{2} = \mathbf{x}^T\mathbf{Mx}$, is the summation of all cross terms, each of which is weighted by the corresponding element in the matrix $\mathbf{M}$. For example, if $n = 3$, the quadratic term is the summation of 9 cross terms:
\begin{align}
    \mathbf{x}^T\mathbf{M}\mathbf{x}    & = \sum_{i=1}^3{\sum_{j=1}^3 \mathbf{x}_i\mathbf{M}_{i,j}\mathbf{x}_j}. 
\end{align}
Due to the inclusion of interactions between any two inputs, the general quadratic neuron exhibits strong non-linearity, yielding better expressivity than the linear neuron.

\subsection{Complexity Reduction of the General Quadratic Neuron}
Although the general quadratic neuron is theoretically superior, it introduces significant additional parameters and computation costs. In Table~\ref{tab:neurons} we provide a comprehensive summary of various quadratic neurons, including the mathematical formulations as well as the associated parameter and computational complexities. For simplicity, we ignore the bias term in the formulation and complexity analysis. There has been attempt to integrate the general quadratic neuron shown in Eq.~\eqref{eq:gen_quad_neuron}, which has a parameter complexity of $n^2+n$, into modern DNNs \cite{zoumpourlis2017non}. To avoid excessive complexity, the general quadratic neurons are only deployed in the initial layer \cite{zoumpourlis2017non}. Another variant of the general quadratic neuron without the linear term has been utilized in \cite{mantini2021cqnn}. Unfortunately, the quadratic complexity still makes it impractical to implement such quadratic neurons in modern neural networks directly.

\begin{table}
\begin{center}
\begin{threeparttable}
\caption{Summary of Quadratic Neurons.}\label{tab:neurons}
\begin{tabular}{|m{0.65cm}|l|m{1.45cm}|m{1.45cm}|}
\hline
\multicolumn{1}{|l|}{} & \multicolumn{1}{l|}{Neuron Format} & \multicolumn{1}{l|}{Parameter} & \multicolumn{1}{l|}{Computation} \\ \hline
\multicolumn{1}{|l|}{\cite{zoumpourlis2017non}} & \multicolumn{1}{l|}{$\mathbf{x}^T\mathbf{M}\mathbf{x} + \mathbf{w}^T\mathbf{x}$} & \multicolumn{1}{l|}{$\mathcal{O}(n^2+n)$} & \multicolumn{1}{l|}{$\mathcal{O}(n^2+2n)$} \\ \hline
\multicolumn{1}{|l|}{\cite{mantini2021cqnn}} & \multicolumn{1}{l|}{$\mathbf{x}^T\mathbf{M}\mathbf{x}$} & \multicolumn{1}{l|}{$\mathcal{O}(n^2)$} & \multicolumn{1}{l|}{$\mathcal{O}(n^2+n)$} \\ \hline
\multicolumn{1}{|l|}{\cite{QuadraticRes}} & \multicolumn{1}{l|}{$(\mathbf{w}_1^T\mathbf{x})(\mathbf{w}_2^T\mathbf{x})+\mathbf{w}_1^T\mathbf{x}$} & \multicolumn{1}{l|}{$\mathcal{O}(2n)$} & \multicolumn{1}{l|}{$\mathcal{O}(2n)$} \\ \hline
\multicolumn{1}{|l|}{\cite{jiang2020nonlinear}} & \multicolumn{1}{l|}{$\mathbf{x}^T\mathbf{Q}_1^k (\mathbf{Q}_2^k)^T\mathbf{x} + \mathbf{w}^T\mathbf{x}$} & \multicolumn{1}{l|}{$\mathcal{O}(2kn+n)$} & \multicolumn{1}{l|}{$\mathcal{O}(2kn+k)$} \\ \hline
\multicolumn{1}{|l|}{\cite{fan2021expressivity}} & \multicolumn{1}{l|}{$(\mathbf{w}_1^T\mathbf{x})(\mathbf{w}_2^T\mathbf{x})+\mathbf{w}_3^T(\mathbf{x}^{\odot 2})$} & \multicolumn{1}{l|}{$\mathcal{O}(3n)$} & \multicolumn{1}{l|}{$\mathcal{O}(4n)$} \\ \hline
\multicolumn{1}{|l|}{\cite{QuadraLib_xu}} & \multicolumn{1}{l|}{$(\mathbf{w}_1^T\mathbf{x})(\mathbf{w}_2^T\mathbf{x})+\mathbf{w}_3^T\mathbf{x}$} & \multicolumn{1}{l|}{$\mathcal{O}(3n)$} & \multicolumn{1}{l|}{$\mathcal{O}(3n)$} \\ \hline
\multicolumn{1}{|l|}{\multirow{2}{*}{Ours}} & \multicolumn{1}{l|}{\multirow{2}{*}{\begin{tabular}[c]{@{}l@{}}$\{\mathbf{x}^T \mathbf{Q}^k \mathbf{\Lambda}^k (\mathbf{Q}^k)^T\mathbf{x} $ \\ $+ \mathbf{w}^T\mathbf{x},\  \mathbf{x}^T \mathbf{Q}^k\}$\end{tabular}}} & \multicolumn{1}{l|}{\multirow{2}{*}{$\mathcal{O}(n+\frac{k}{k+1})$}} & \multicolumn{1}{l|}{\multirow{2}{*}{$\mathcal{O}(n+\frac{2k}{k+1})$}} \\
\multicolumn{1}{|l|}{} & \multicolumn{1}{l|}{} & \multicolumn{1}{l|}{} & \multicolumn{1}{l|}{} \\ \hline
\end{tabular}
   \begin{tablenotes}[para,flushleft]\footnotesize
    \item[1] Subscripts in this table are used to distinguish different items.\\
    \item[2] $\mathbf{x}, \mathbf{w}\in \mathbb{R}^n,\mathbf{M}\in \mathbb{R}^{n\times n}$, where $n$ is the number of neuron's inputs. \\
    \item[3] $\mathbf{Q}^k \in \mathbb{R}^{n\times k}, \mathbf{\Lambda}_k \in \mathbb{R}^{k\times k}$, where $k$ is the rank of decomposition. \\
    \item[4] $(\cdot)^{\odot 2}$ is the element-wise square operation. \\
    \item[5] $\{,\}$ is the concatenation operation.
    \end{tablenotes}
\end{threeparttable}
\end{center}
\end{table}

Recently, some attention has been given to simplified quadratic neurons that can maintain the non-linearity of the general quadratic neuron without incurring excessive parameter overhead \cite{jiang2020nonlinear,QuadraticRes,fan2021expressivity,QuadraLib_xu}. One approach to address the complexity issues associated with quadratic neurons is to employ low-rank decomposition. In \cite{jiang2020nonlinear}, the general quadratic form is simplified with low-rank decomposition, resulting in a reduction of complexity from $n^2$ to $2kn$, where $k$ is the rank of decomposition ranging from 1 to $n$. \cite{QuadraLib_xu, fan2021expressivity} has proposed a neuron that is similar to \cite{jiang2020nonlinear} but with a fixed rank of 1.

Our proposed quadratic neuron is presented in the last row of Table~\ref{tab:neurons}. This neuron also incorporates a hyper-parameter $k$, which is the rank of decomposition. However, unlike prior work such as \cite{jiang2020nonlinear}, the parameter and computational complexities of our design do not increase proportionally to $k$. This property allows for more flexible choices of different decomposition ranks.

By investigating previous studies, we summarize the main challenges associated with quadratic neurons as follows:

\begin{itemize}
    \item \textbf{Extra Parameter and Computational Cost}: Despite the efforts to simplify the general quadratic form to reduce its complexity to linear levels \cite{jiang2020nonlinear, fan2021expressivity, QuadraLib_xu}, there still exists considerable overhead compared to the linear neuron. Some work \cite{jiang2020nonlinear, zoumpourlis2017non} has attempted to mitigate these issues by only applying the complex neurons to the initial or some selected layers. However, this partial deployment impairs the deployment flexibility of these neurons.

    \item \textbf{Compromised Expressivity}: While some existing methods have leveraged simplified versions of general quadratic neurons, many of them have sacrificed the expressivity to reduce the neuron complexity \cite{QuadraLib_xu, fan2021expressivity, QuadraticRes}. For example, the quadratic neuron in \cite{jiang2020nonlinear} allows the adjustment of expressivity with a hyper-parameter $k$. However, as shown in Table~\ref{tab:neurons}, the cost is proportional to the value of $k$, making it impractical to increase the neuron's expressivity with a large $k$. Notably, only neurons with low ranks such as $k=1,2$ and $3$ have been explored in \cite{jiang2020nonlinear}.

    \item \textbf{Underutilization of Internal Features}: Though quadratic neurons introduce complex computations to the inputs, the computational results are simply summed to a singular value, which forms the neuron's output. The direct summation of all intermediate terms may obscure the features extracted by different computational processes instead of exhaustively making use of their information. 

\end{itemize}

\section{Proposed Efficient Quadratic Neuron} \label{sec:method}

In this section, we present our proposed novel quadratic neuron as a solution to the aforementioned issues in Sec.~\ref{sec:motivation}. To maintain the neuron's expressivity, we follow a strict procedure to simplify the general quadratic neuron. Additionally, by incorporating the reuse of intermediate information within the neuron, the required number of neurons can be further reduced for the same size of feature maps, thereby lowering computational and storage complexities.

\subsection{Quadratic Matrix Decomposition}\label{subsec:neuron_simplify}

As previously noted in Sec.~\ref{sec:motivation}, the general quadratic neuron consists of two terms: the linear term and the quadratic term. In this section, we mainly discuss the quadratic term which comprises the majority of parameters.

Compared with previous methods which have fixed and compromised expressivity \cite{fan2021expressivity,QuadraLib_xu,QuadraticRes} or do not fully exploit the characteristics of the quadratic neuron, such as symmetry \cite{jiang2020nonlinear}, we adopt a systematic approach that utilizes the symmetry of the quadratic matrix. Starting from the general quadratic neuron, we manage to reduce the number of parameters in the quadratic neuron while maintaining its expressive ability to the fullest extent possible.

To facilitate the subsequent construction, we present the following lemma to illustrate the property of quadratic neurons:

\newtheorem{Lemma}{Lemma}
\begin{Lemma}\label{lemma:symmetric}
For any real-valued $n\times n$ matrix $\mathbf{M}$, there exists a real-valued \textbf{symmetric} matrix $\mathbf{M}'$ such that:
\begin{equation}
\label{eq:lemma_1}
    \mathbf{x}^T\mathbf{M}\mathbf{x} = \mathbf{x}^T\mathbf{M}'\mathbf{x}, \forall \mathbf{x}\in \mathbb{R}^n.
\end{equation}
\end{Lemma}

\begin{proof}
To prove this, we can split the quadratic term into two halves and transpose one of them:
\begin{align}
     \mathbf{x}^T\mathbf{M}\mathbf{x} 
     & = \frac{1}{2} \left(\mathbf{x}^T\mathbf{M}\mathbf{x} + \left(\mathbf{x}^T\mathbf{M}\mathbf{x}\right)^T\right) \notag\\
     & = \mathbf{x}^T\left(\frac{\mathbf{M}+\mathbf{M}^T}{2}\right)\mathbf{x}.
\end{align} By defining $\mathbf{M}' \triangleq (\mathbf{M}+\mathbf{M}^T)/2$, which is a real-valued symmetric matrix, we prove the lemma.
\end{proof}

With Lemma~\ref{lemma:symmetric}, the matrix $\mathbf{M}$ can always be replaced by a symmetric matrix without affecting the output. Therefore, for simplicity in notation, we assume that the matrix $\mathbf{M}$ is already symmetric in the following discussions.

To simplify the quadratic neuron, we use a low-rank decomposition method which is similar to \cite{jiang2020nonlinear}. However, there are two key differences in our methodology. Firstly, we utilize eigendecomposition to factorize $\mathbf{M}$ into three matrices. Secondly, we leverage the matrix symmetry from Lemma~\ref{lemma:symmetric} to further reduce the complexity by half compared with \cite{jiang2020nonlinear}.

According to the spectral theorem in linear algebra, which states that a real symmetric matrix can be diagonalized using eigendecomposition with an orthonormal matrix, we simplify the quadratic matrix as follows:
\begin{equation}
    \mathbf{M} = \mathbf{Q}\mathbf{\Lambda}\mathbf{Q}^{T},
\end{equation}
where $\mathbf{Q}$ is an orthonormal matrix and $\mathbf{\Lambda}$ is a diagonal matrix.

\begin{figure}[tbp]
\centerline{\includegraphics[width=2.5in]{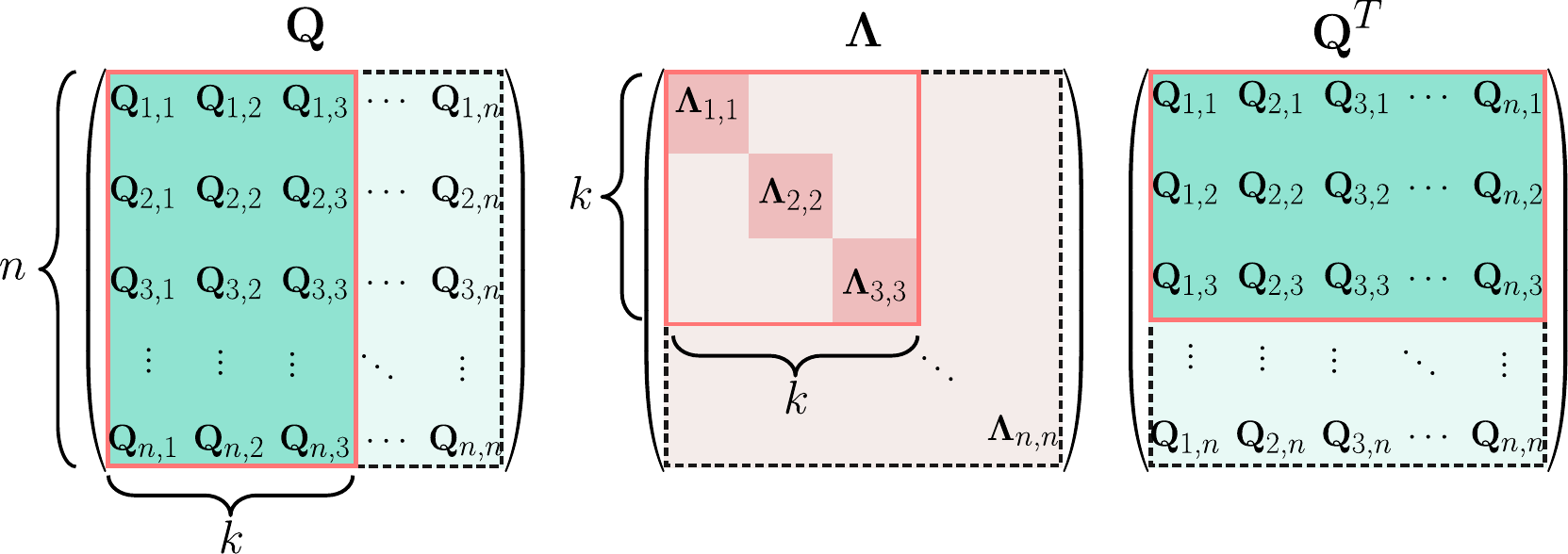}}
\caption{Top-$k$ selection of the decomposed matrices $\mathbf{Q}$ and $\mathbf{\Lambda}$.} \label{fig:quad_prune}
\end{figure}

The eigendecomposition possesses a crucial property that the order of eigenvalues in matrix $\mathbf{M}$ may be interchanged, if the corresponding eigenvectors are also swapped in the same order. This property is leveraged in principal component analysis (PCA) by sorting the eigenvalues in a descending order based on their magnitudes \cite{doi:https://doi.org/10.1002/0470013192.bsa501}. For simplicity in notation, we make the assumption that the elements of matrix $\mathbf{\Lambda}$ have already been sorted in descending order.

In this study, we adopt a method that retains the top-$k$ eigenvalues in terms of their magnitude. This method achieves the optimal rank-$k$ approximation for a matrix in terms of the Frobenius norm according to the Eckart-Young-Mirsky theorem \cite{eckart1936approximation}. Fig.~\ref{fig:quad_prune} illustrates the top-$k$ selection from the original matrices, resulting in an approximation of $\mathbf{M}$ denoted as $\mathbf{M}^k$. The approximation can be formulated as:
\begin{equation}
    \mathbf{M}  \approx  \mathbf{M}^k = \mathbf{Q}^k\mathbf{\Lambda}^k({\mathbf{Q}^k})^{T},
\end{equation}
where $\mathbf{Q}^k$ is a $n\times k$ matrix pruned from the first $k$ columns of $\mathbf{Q}$, and $\mathbf{\Lambda}^k$ is a $k\times k$ matrix with top $k$ eigenvalues placed at its diagonal.

We denote the quadratic term with rank-$k$ approximation as:
\begin{align}
\label{eq:quad_decomp_k}
    y_{2} \approx y_2^k = \mathbf{x}^T \mathbf{Q}^k\mathbf{\Lambda}^k({\mathbf{Q}^k})^{T} \mathbf{x}. 
\end{align}

With the use of inherent symmetry in the quadratic matrix, our approximation can reduce the complexity in terms of the number of parameters from $n^2$ to $kn+k$, where $k$ is an adjustable hyper-parameter, allowing us to control the neuron's expressivity. Compared with previous works using low-rank approximation \cite{jiang2020nonlinear}, our method reduces the complexity significantly while fully preserving the expressive ability.

\begin{figure}[tbp]
\centerline{\includegraphics[width=2.5in]{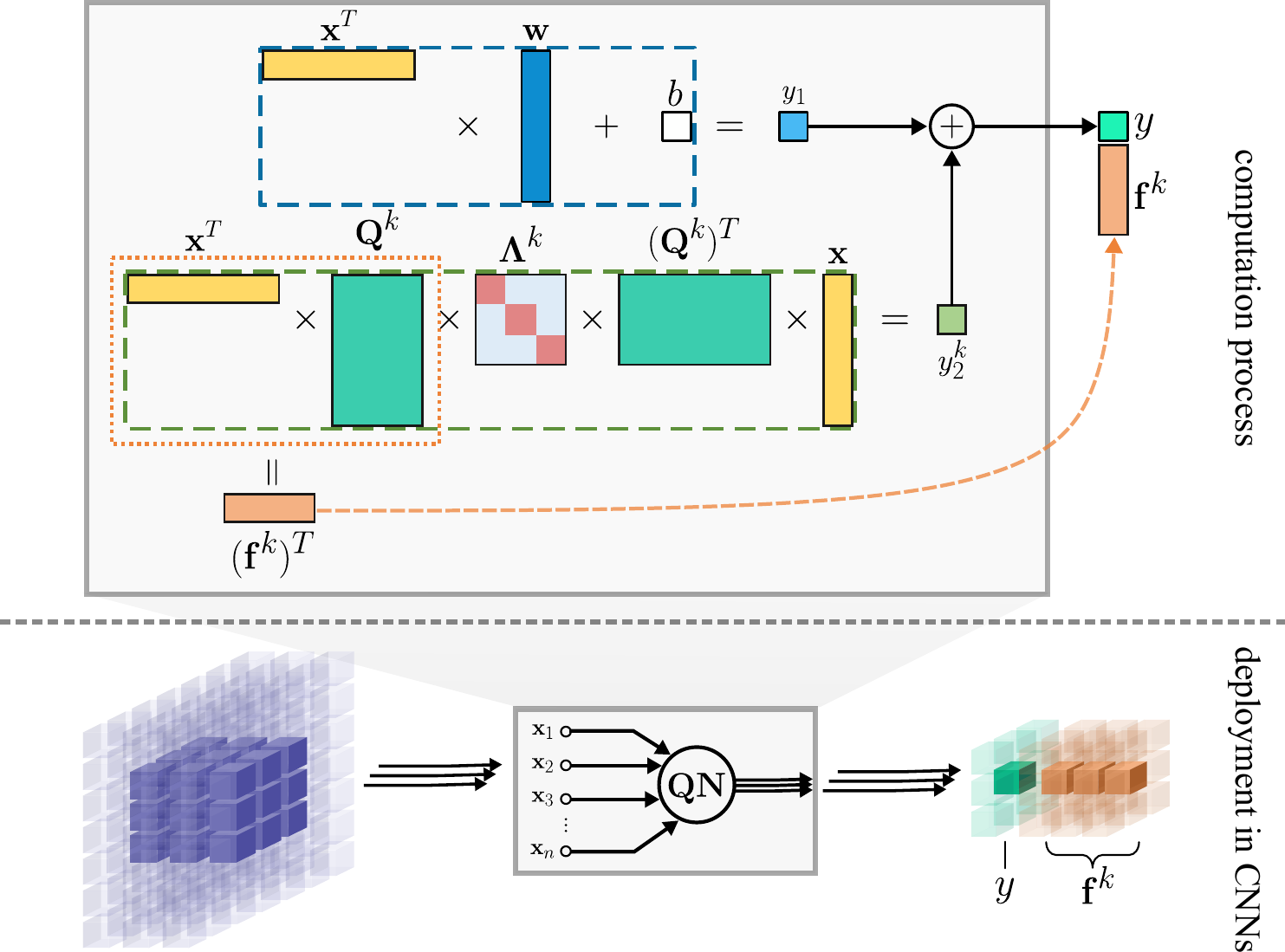}}
\caption{Computation process of the proposed quadratic neuron and its deployment in a convolutional layer.} \label{fig:our_neuron_full}
\end{figure}

\subsection{Information Utilization in Quadratic Neurons}\label{subsec:vectorized_output}

Complex neurons in prior research try to improve the expressive ability of a neural network by introducing additional parameters. Consequently, these modifications lead to the generation of more intermediate results during the computation, including the cross terms. However, most of them simply collapse multiple terms into a singular output using additions. Such a simplistic method of reduction can result in a significant loss of information, particularly when the number of intermediate results is substantial. For example, the general quadratic neuron generates $n^2$ cross terms, each of which may represent some kind of feature. The simple summation may inhibit the expression of these features, causing the underutilization of internal information in a neuron.

To address this issue, we propose a scheme that outputs the intermediate features of a quadratic neuron to the next layer. It can be observed that the computation process of $({\mathbf{Q}^k})^{T}\mathbf{x}$ in Eq.~\eqref{eq:quad_decomp_k} is mathematically equivalent to $k$ linear neurons with each column of $\mathbf{Q}^k$ serving as the linear weights. By defining the intermediate product as:
\begin{equation}
    \mathbf{f}^k \triangleq ({\mathbf{Q}^k})^{T}\mathbf{x},
\end{equation}
where $\mathbf{f}^k$ is a vector with the length of $k$, each element of $\mathbf{f}^k$ can be viewed as the output of a linear neuron. These intermediate features may also contain useful information which can be used by the following layers. Therefore, to achieve exhaustive utilization of this internal information, we take $\mathbf{f}^k$ as additional outputs besides the original output $y$, which is the sum of $y_1$ and $y_2^k$, to the next layer, resulting in a neuron with a vectorized output. Fig.~\ref{fig:our_neuron_full} illustrates the structure of the proposed quadratic neuron, where $y$ and $\mathbf{f}^k$ are concatenated as the final output, and its deployment in a CNN. When deployed in a convolutional layer of a CNN, the additional outputs are placed along the channel dimension, which means that each convolutional filter can output multiple channels.

\subsection{Parameter and Computational Complexity Analysis}\label{subsec:neuron_complex_analysis}
In this section, we discuss the complexity of our proposed neuron in terms of \textbf{storage cost} (number of \textit{parameters}) and \textbf{computational cost} (number of \textit{multiply-accumulate operations} (MACs)). For simplicity, we ignore the cost of the bias term, which is negligible compared to other operations.

\textbf{Parameter}: Each quadratic neuron contains the following trainable parameters: $\mathbf{Q}^k\in\mathbb{R}^{n\times k}$, $\mathbf{\Lambda}^k\in\mathbb{R}^{k\times k}$, and $\mathbf{w}\in\mathbb{R}^{n\times 1}$. Note that $\mathbf{\Lambda}^k$ is a diagonal matrix thus it has only $k$ parameters. The total number of parameters is the sum of all mentioned terms above:
\begin{equation}\label{eq:param_cost}
    \#\texttt{Parameter} = (k+1)n+k.
\end{equation}

\textbf{Computation}: The computation process of our proposed neuron contains two parts as shown in Fig.~\ref{fig:our_neuron_full}: the linear part $\mathbf{w}^T\mathbf{x}$ with the cost of $n$, and the quadratic part $\mathbf{x}^T \mathbf{Q}^k\mathbf{\Lambda}^k({\mathbf{Q}^k})^{T} \mathbf{x}$ in Eq.~\eqref{eq:quad_decomp_k} which consists of four matrix multiplications.

The overall computational cost of the quadratic part depends on the order of matrix multiplications. We first compute the product of $(\mathbf{Q}^k)^T$ and $\mathbf{x}$. Then the next step is to compute $y_2^k=(\mathbf{f}^k)^T\mathbf{\Lambda}^k\mathbf{f}^k$. The computation cost of $\mathbf{f}^k = (\mathbf{Q}^k)^T\mathbf{x}$ is $kn$, and the computation of $(\mathbf{f}^k)^T\mathbf{\Lambda}^k$ which contains $k$ element-wise multiplications has a cost of $k$. $y_2^k=(\mathbf{f}^k)^T\mathbf{\Lambda}^k\mathbf{f}^k$, which requires twice such multiplications, has a cost of $2k$. Accordingly, the overall cost  of our proposed neuron structure is the sum of $n$ from the linear part, $kn$ from $\mathbf{f}^k = (\mathbf{Q}^k)^T\mathbf{x}$, and $2k$ from $y_2^k=(\mathbf{f}^k)^T\mathbf{\Lambda}^k\mathbf{f}^k$:
\begin{equation}\label{eq:compute_cost}
    \#\texttt{MAC} = (k+1)n+2k.
\end{equation}

\textbf{Averaged Complexity}: As we mentioned in Sec.~\ref{subsec:vectorized_output}, each of our neurons generates $k+1$ outputs. Therefore, when deploying them to a DNN, fewer neurons are required to obtain the original sizes of feature maps. To reflect the cost of our neurons in practice more accurately, we analyze the complexity of a neuron averaged by the size of its output. With Eqs.~\eqref{eq:param_cost} and \eqref{eq:compute_cost}, the averaged parameter and computation cost of our proposed neuron per output are $n+\frac{k}{k+1}$ and $n+\frac{2k}{k+1}$, respectively. Considering that $n$ is large in DNNs, the additional overhead is negligible compared with the linear neurons.
\section{Experimental Results}\label{sec:experiment}

This section presents a series of experiments that evaluate the efficacy of our proposed neuron when deployed in existing DNNs. In Sec.~\ref{subsec:network_deploy} and \ref{subsec:transformer}, we assess the impact of our neuron on the accuracy as well as computational and parameter costs of DNNs in image classification tasks and WMT14 English-German translation task. In Sec.~\ref{subsec:quad_explore}, we analyze the proposed quadratic neuron, including parameter distributions, and visualization of the neuron's response in CNNs. 

 
\subsection{Image Classification}\label{subsec:network_deploy}
Within this subsection, we assess the effectiveness of our proposed neuron by implementing it across all convolutional layers in a variety of CNNs. Our evaluation consists of analyzing the accuracy of the resulting CNNs in image classification tasks. In addition, we also conduct an analysis of the associated costs of parameters and computations.

\begin{figure}[t]
\centerline{\includegraphics[width=3.3in]{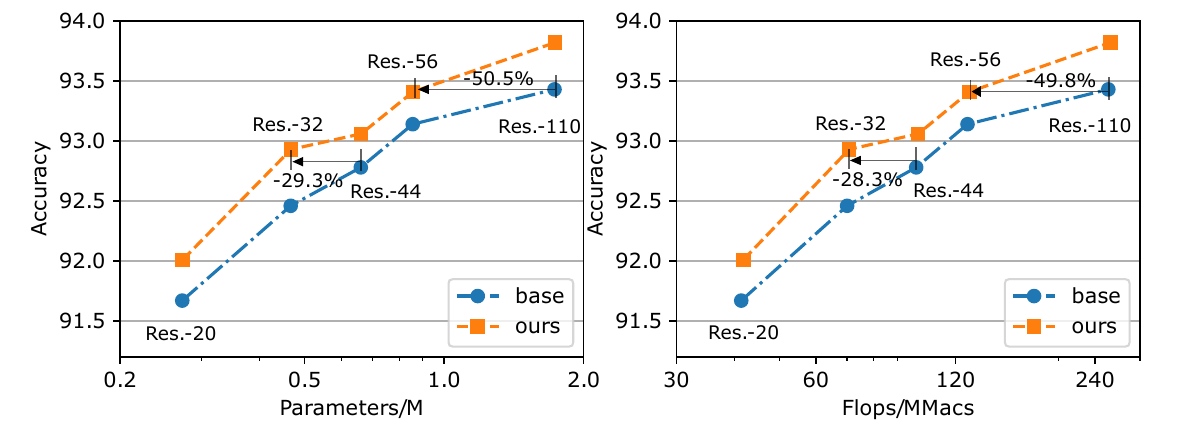}}
\caption{Accuracy on CIFAR-10, number of parameters and FLOPs of ResNets (20, 32, 44, 56 and 110) with linear and quadratic neurons.} \label{fig:base_eff}
\end{figure}

\subsubsection{Evaluation on Efficiency}
The CIFAR-10 dataset \cite{CIFAR_dataset} consists of 60$k$ $32\times 32$ images categorized into 10 classes. Prior to training, we applied a random crop to each image to $32\times32$ from $40\times 40$ images with paddings, as well as a random horizontal flip as data augmentation. All networks were trained with SGD optimizer for 180 epochs with a training batch size of 128. The learning rate was initially set to 0.1 and was decayed by a factor of 0.1 at epochs 90 and 135. The learning rate of $\mathbf{\Lambda}^k$ in our proposed neuron was initially set to $10^{-4}$ and the rank of decomposition $k$ was fixed to 9.

As shown in Fig~\ref{fig:base_eff}, ResNets with our proposed neurons can achieve significant improvement in terms of accuracy and the computational and storage efficiency compared to the baseline networks. For example, ResNet-32 with the proposed quadratic neurons can achieve better accuracy than ResNet-44 with conventional linear neurons with 29.3\% fewer parameters and 28.3\% less computational cost. The improvement is even more significant for deeper networks: ResNet-56 with quadratic neurons can reduce the computational cost and storage cost approximately by 50\% while having a similar accuracy with baseline ResNet-110. 

We also compared our proposed neuron with previous quadratic neurons. To ensure a fair comparison, we expanded the networks equipped with our proposed neurons by adding channels to provide a slight accuracy advantage. This enabled us to compare different neurons based on their parameter and computation costs. We reproduced the results of \cite{QuadraLib_xu} using the same configuration as ours, while the results of \cite{fan2021expressivity} were obtained from the original paper. 


\begin{figure}[t]
\centerline{\includegraphics[width=3.2in]{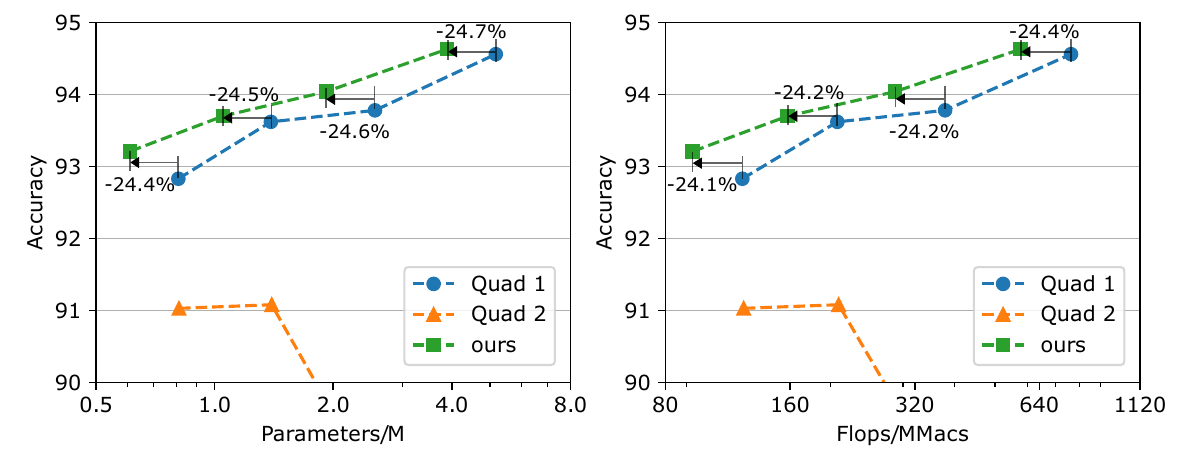}}
\caption{Accuracy on CIFAR-10, number of parameters and FLOPs of ResNets (20, 32, 56 and 110) with quadratic neurons from \cite{fan2021expressivity} (Quad 1), \cite{xu2020efficient} (Quad 2), and this work.} \label{fig:quad_eff}
\end{figure}

Fig~\ref{fig:quad_eff} presents a comparison of the accuracy, parameter, and computational costs of networks with quadratic neurons from previous research \cite{fan2021expressivity,QuadraLib_xu} and this work. Among these designs, our proposed neuron structure demonstrates the best efficiency in terms of storage and computations. Notably, as the network depth increases, the design from \cite{QuadraLib_xu} performs worse on the test set, with the accuracy falling below $90\%$. Compared to \cite{fan2021expressivity}, our proposed design is able to achieve significant better accuracy with at least 24.4\% less storage cost and 24.1\% less computational cost.

\subsubsection{Training Stability}

Kervolutional neuron (KNN) \cite{kervolutional_nn} is a non-linear design without any additional parameters. In this section, we demonstrate that, despite the advantages of KNN in the number of parameters, the proposed quadratic neuron can guarantee a more stable training process and results in better accuracy. We trained ResNet-18 with different configurations on the ILSVRC-2012 dataset \cite{ILSVRC15}. The networks were trained for 100 epochs with a batch size of 256. The learning rate was set 0.1, and was decayed by 0.1 at epochs 30, 60, and 90. Regarding our proposed neuron, the learning rate of $\mathbf{\Lambda}^k$ was set to $10^{-5}$, and the rank of decomposition $k$ was fixed to 9. We equipped the first $n$ convolutional layers with neurons from \cite{kervolutional_nn} (referred to as ``KNN-$n$''), as well as deploying our proposed neurons to all convolutional layers for comparison. 

\begin{figure}[t]
\centerline{\includegraphics[width=3.2in]{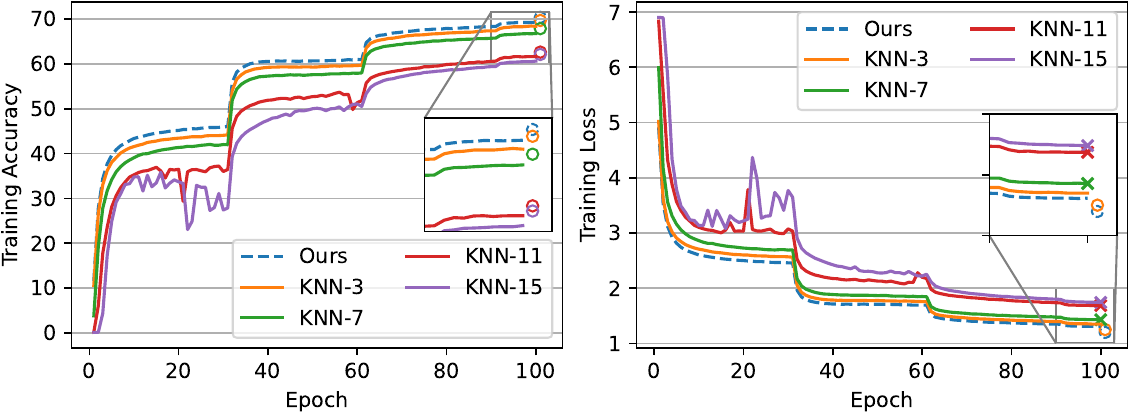}}
\caption{Training stability analysis of ResNet-18 with KNN \cite{kervolutional_nn} and the proposed neurons on ImageNet. "KNN-$n$" refers to the network equipped with neurons from \cite{kervolutional_nn} in first $n$ layers.} \label{fig:knn_stab}
\end{figure}

The training accuracy and loss over time is shown in Fig.~\ref{fig:knn_stab}, where the circles represent the validation accuracy and loss after training, and a cross mark indicates that the loss does not converge on the validation set. As can be found in Fig.~\ref{fig:knn_stab}, when only deployed in the first 3 layers, \cite{kervolutional_nn} can achieve stable training and similar accuracy to our proposed design. However, as with more deployed layers such as 11 and 15, the training process becomes unstable as obvious fluctuation can be found, resulting in bad performance in the end. Also, extreme values can be found during the testing process, indicating the inference on unseen test data is also non-stable. In summary, compared to \cite{kervolutional_nn}, our proposed neuron can achieve stable training with the highest accuracy while being applied to all layers. This brings superior flexibility for the deployment of our neuron.

\subsection{Machine Translation with Transformers}\label{subsec:transformer}
In this section, we deploy the proposed quadratic neuron into the Transformer architecture, including all linear projection operators in the multi-head attention blocks. The experiments followed the same settings as in \cite{transformer} unless otherwise stated. The network performance was evaluated on the English-to-German newstest2014 dataset and we used BLEU scores to evaluated the translation quality (higher BLEU score is better). Table~\ref{tab:transformer_eff} shows the BLEU scores in the translation task with different evaluation settings. Since the number of FLOPs averaged by the input length of a Transformer is approximately twice the number of parameters \cite{scaling_law_llm}, we only report the number of network parameters in Table~\ref{tab:transformer_eff}. The last three columns present the results of networks with the proposed quadratic neurons with different learning rate configurations ($10^{-4}, 10^{-5}$ and $10^{-6}$) to quadratic parameters $\Lambda^k$. Compared with the baseline, the quadratic transformer can achieve better performance with significantly fewer ($20.3\%$) parameters and FLOPs.

\begin{table}[t]
\begin{center}
\begin{threeparttable}
\caption{Performance on English-to-German newstest2014 dataset and parameter cost of Transformers.}
\label{tab:transformer_eff}
\begin{tabular}{|ccc|c|rrr|}
\hline
\multicolumn{1}{|c|}{}                      & \multicolumn{2}{c|}{Eval. Setting}          & \multicolumn{1}{c|}{Baseline}                                 & \multicolumn{3}{c|}{Quadratic}                                                    \\ \cline{2-7}
\multicolumn{1}{|c|}{} & \multicolumn{1}{c|}{Tokenization}  & Cased                    & \multicolumn{1}{c|}{-} & \multicolumn{1}{c|}{1E-4} & \multicolumn{1}{c|}{1E-5} & \multicolumn{1}{c|}{1E-6} \\ \hline 
\multicolumn{1}{|c|}{\multirow{4}{*}{BLEU\footnotemark[1]}}                      & \multicolumn{1}{c|}{13a}           & Yes           & 25.3                     & \multicolumn{1}{r|}{25.2} & \multicolumn{1}{r|}{25.2} & \textbf{25.4}             \\ \cline{2-7} 
\multicolumn{1}{|c|}{}                      & \multicolumn{1}{c|}{13a}           & No            & 25.7                     & \multicolumn{1}{r|}{25.7} & \multicolumn{1}{r|}{25.7} & \textbf{25.9}             \\ \cline{2-7} 
\multicolumn{1}{|c|}{}                      & \multicolumn{1}{c|}{International} & Yes           & 26.0                     & \multicolumn{1}{r|}{25.9} & \multicolumn{1}{r|}{25.9} & \textbf{26.1}             \\ \cline{2-7} 
\multicolumn{1}{|c|}{}                      & \multicolumn{1}{c|}{International} & No            & 26.4                     & \multicolumn{1}{r|}{26.3} & \multicolumn{1}{r|}{26.4} & \textbf{26.6}             \\ \hline
\multicolumn{3}{|c|}{\#Params/M}                                                & 15.7                     & \multicolumn{3}{c|}{\textbf{12.6} (-20.3\%)}                                      \\ \hline

\end{tabular}
   \begin{tablenotes}[para,flushleft]\footnotesize
    \item[1] Higher is better.
    \end{tablenotes}
\end{threeparttable}
\end{center}
\end{table}

\subsection{Analysis of the Quadratic Neuron}\label{subsec:quad_explore}
In this section, to comprehensively analyze the proposed quadratic neuron, experiments of the ResNet-20 network on the CIFAR-100 dataset \cite{CIFAR_dataset} are conducted. The networks are trained with a batch size of 64 for 250 epochs.

\subsubsection{Parameter Distribution}

\begin{figure}[t]
\centerline{\includegraphics[width=3.6in]{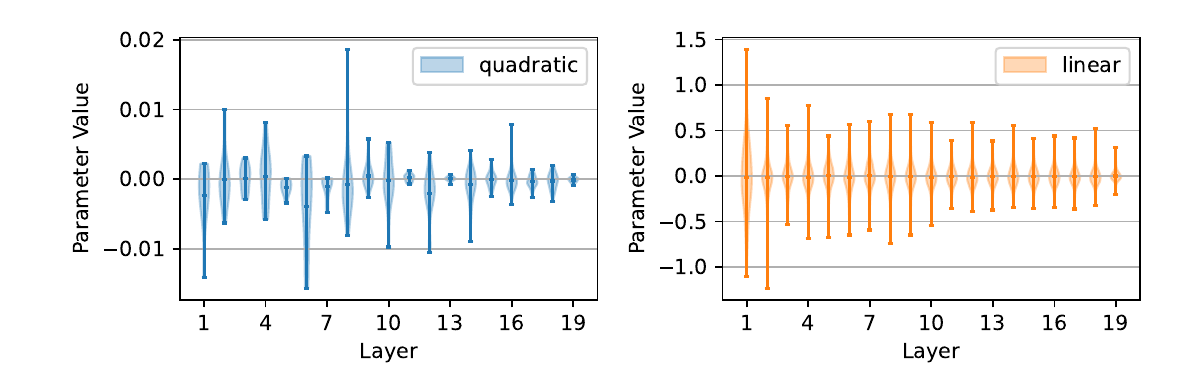}}
\caption{Distribution of linear and quadratic parameters in ResNet-20 after training on CIFAR-100.} \label{fig:params_vs_layer}
\end{figure}

Fig.~\ref{fig:params_vs_layer} presents the distribution of linear and quadratic parameters in different layers of ResNet-20 after the training process. It can be observed that, in contrast to linear parameters, the distribution range of quadratic parameters has a larger variance with respect to depth. Specifically, the values of quadratic parameters are more significant in layers 1, 6, and 8, while in some layers such as 11, 13, and 19, the quadratic parameters tend to be more centered around zero. This observation suggests that, unlike linear neurons, quadratic neurons may not be essential for every layer. However, our findings also suggest that the application of high-order neurons only in the first layer \cite{kervolutional_nn,zoumpourlis2017non} may not be the optimal choice.

\subsubsection{Neuron Response Analysis}
\begin{figure}[t]
\centerline{\includegraphics[width=2.5in]{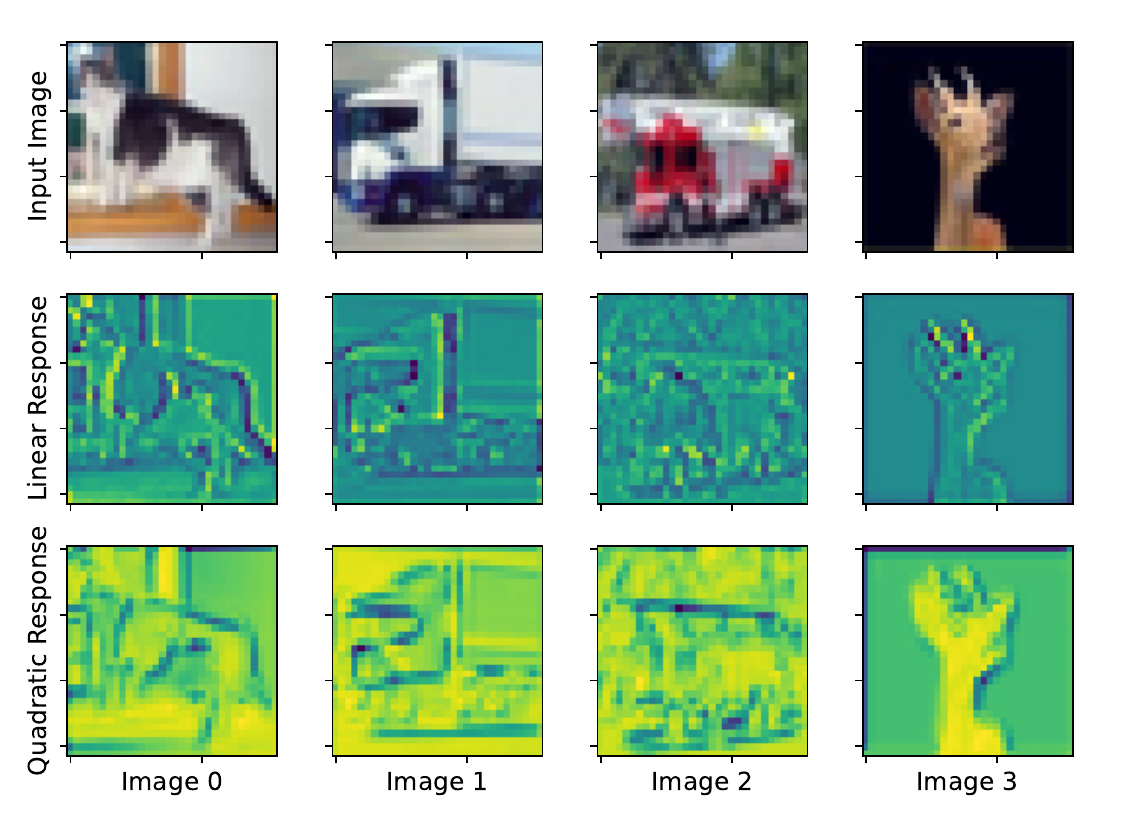}}
\vspace{-0.2cm}
\caption{Response visualization of linear and quadratic neurons.}\label{fig:neural_response}
\end{figure}

To gain further insights into the functionality of quadratic neurons in CNNs, we conducted a visualization experiment to examine their outputs. Fig.~\ref{fig:neural_response} illustrates the responses of both linear and quadratic parts of our neurons, where the linear and quadratic responses correspond to $\mathbf{w}^T\mathbf{x}
+b$ and $y_2^k$, respectively. Our analysis indicates that, in contrast to linear parts that tend to extract edges, quadratic neurons focus more on the entire object. This property is particularly evident in the deer image, where the whole shape is clearly distinguished from the background. We also discover that quadratic neurons tend to extract low-frequency information such as the shape of an object, while high-frequency details such as texture or noise are less significant in the quadratic response.

\section{Conclusion}\label{sec:conclusion}
In this paper, we introduce an efficient quadratic neuron for DNNs using a systematic approximation approach and the vectorized-output technique. Experimental results demonstrate its superior computational and storage efficiency, training stability, and deploy-flexibility compared to prior non-linear neuron designs.

\let\oldbibliography\thebibliography
\renewcommand{\thebibliography}[1]{%
\oldbibliography{#1}%
\fontsize{7.2pt}{7.2}\selectfont
\setlength{\itemsep}{0.1pt}%
}

\bibliographystyle{IEEEtran}
\bibliography{bib}

\end{document}